\documentclass[10pt,twocolumn,letterpaper]{article}

\usepackage{wacv}
\usepackage{times}
\usepackage{epsfig}
\usepackage{graphicx}
\usepackage{amsmath}
\usepackage{amssymb}
\usepackage{color}
\newcommand{\changed}[1]{\textcolor{black}{#1}}
\newcommand{\CRC}[1]{\textcolor{black}{#1}}

\wacvfinalcopy 

\def\wacvPaperID{472} 

\ifwacvfinal\fi
\begin{document}

\title{Characteristic Regularisation for Super-Resolving Face Images}

\author{Zhiyi Cheng \\
Queen Mary University of London\\
{\tt\small z.cheng@qmul.ac.uk}
\and
Xiatian Zhu \\
Vision Semantics Limited, London, UK\\
{\tt\small eddy.zhuxt@gmail.com}
\and
Shaogang Gong \\
Queen Mary University of London \\
{\tt\small s.gong@qmul.ac.uk}
}

\maketitle
\ifwacvfinal\thispagestyle{empty}\fi

\begin{abstract}
   Existing facial image super-resolution (SR) methods
	focus mostly on improving ``artificially down-sampled'' low-resolution (LR)
	imagery. 
	Such SR models, although strong at handling artificial LR images,
	often suffer from significant performance drop 
	on genuine LR test data. 
	Previous unsupervised domain adaptation (UDA) methods address this issue
	by training a model 
	using unpaired genuine LR and HR data
	as well as cycle consistency loss formulation.
	However, this renders the model overstretched with two tasks: 
	consistifying the visual characteristics
	and enhancing the image resolution.
	Importantly, this makes the end-to-end model training ineffective 
	due to the difficulty of back-propagating gradients through two concatenated CNNs.
	To solve this problem, we formulate a method
	that joins the advantages of conventional SR and UDA models.
	Specifically, we separate and control the optimisations for characteristics consistifying
	and image super-resolving by introducing Characteristic
        Regularisation (CR) between them.
	This task split makes the model training more effective and 
	computationally tractable.
	Extensive evaluations demonstrate the performance superiority of our method over 
	state-of-the-art SR and UDA models 
	on both genuine and artificial LR facial imagery data.
\end{abstract}

\section{Introduction}

\begin{figure} 
	\centering
	\includegraphics[width=0.85\linewidth]{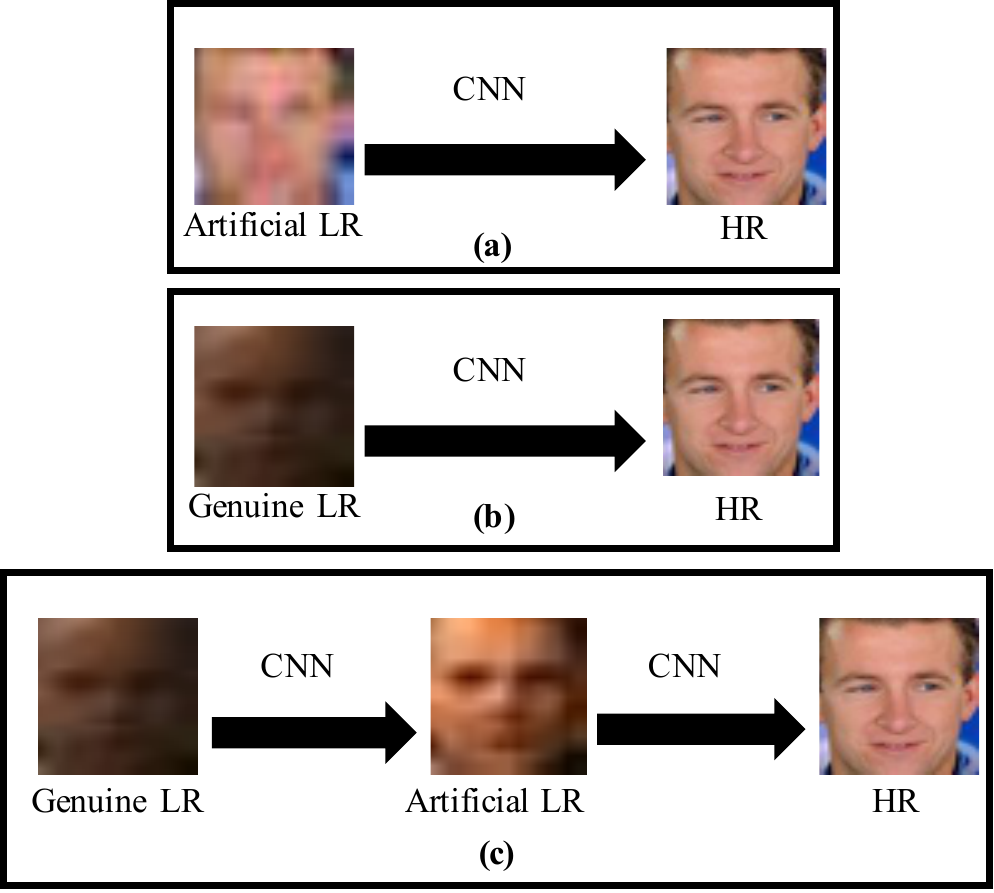}
	\caption{
		CNN architectures for facial image super-resolution.
		{\bf(a)} A CNN is trained to super-resolve {\em artificial} LR facial images
		that are produced by down-sampling \cite{dong2016image,ledig2017photo}.
		It is a supervised learning method.
		{\bf(b)} A CNN learns to adapt a {\em genuine} LR facial images into the HR style. 
		Without LR-HR pairing supervision, a cycle consistency based loss function is often
		used for model training \cite{zhu2017unpaired,yi2017dualgan,bulat2018learn}.
		{\bf(c)} The proposed characteristic regularisation method. 
		The whole model training is regularised by characteristic consistifying
		from genuine LR facial images to artificial LR ones
		before super-resolved to the HR output. 
		Best viewed in colour.
	} 
	\label{fig:designs}
\end{figure}

Facial image analysis \cite{taigman2014deepface,cao2014face,bartlett2003real} 
is significant for many computer vision applications
in business, law enforcement, and public security \cite{pan2010survey}.
However, the model performance often degrades significantly when the face image resolution 
is very low.
Face image super-resolution (SR) \cite{baker2000hallucinating} provides a viable solution 
by recovering a high-resolution (HR) face image from its low-resolution (LR) counterpart.
Existing state-of-the-art image SR models \cite{chen2018fsrnet,yu2018super,zhu2016deep}
mostly learn the low-to-high resolution mapping 
from paired {\em artificial LR} and HR images.
The {artificial} LR images are usually generated by
down-sampling the HR counterparts (Fig \ref{fig:designs}(a)). 
With this paradigm,
existing supervised deep learning models (e.g. CNNs) can be readily applied.
However, this is at a price of poor model generalisation
to {real-world} {\em genuine LR} facial images,
e.g. surveillance imagery captured
in poor circumstances. 
This is because genuine LR data have rather different {\em imaging characteristics} 
from artificial LR images,
often coming with {\em additional}
unconstrained motion blur, noise, corruption, and image compression artefacts. 
(Fig.~\ref{fig:dataset}).
This causes the distribution discrepancy
between training data (artificial LR imagery) and 
test data (genuine LR imagery) which attributes to poor model generalisation,
also known as the domain shift problem \cite{pan2010survey}.
%



Unsupervised domain adaptation (UDA) methods are possible solutions
considering
genuine LR and HR images as two different domains. 
UDA techniques 
have achieved remarkable success
\cite{zhu2017unpaired,hoffman18cycada,taigman2016unsupervised,murez2018image,bousmalis2017unsupervised,yi2017dualgan,kim2017learning,liu2017unsupervised}.
A representative modelling idea is to 
exploit cycle consistency loss functions
between two unpaired domains (Fig \ref{fig:designs}(b))
\cite{zhu2017unpaired,yi2017dualgan,kim2017learning}.
A CNN is used to map an image from one domain
to the other, which is further mapped back by another CNN.
%
With such an encoder-decoder like architecture, 
one can form a reconstruction loss {\em jointly} for both CNN models
{\em without} the need for paired images in each domain. 
The two CNN models can be trained end-to-end, 
inputting an image and outputting a reconstructed image
per domain.
This idea has been attempted in \cite{bulat2018learn}
for super-resolving genuine LR facial imagery.

Using such cycle consistency
for unsupervised domain adaptation has several adverse effects.
The reconstruction loss is applicable only 
to the concatenation of two CNN models.
This exacerbates the already challenging task of
domain adaptation training.
In our context, the genuine LR and HR image domains 
have significant differences in both image resolution and 
imaging conditions.
Compared to a single CNN, the depth of a concatenated CNN-CNN model 
is effectively doubled. 
Existing UDA models apply the cycle consistency loss supervision 
at the final output of the second CNN, and propagate the supervision back 
to the first CNN. This gives rise to extra training difficulties in the form of {\em vanishing} gradients \cite{lee2015deeply,chandar2019towards}.
In addition, jointly training two connected CNN models has to be conducted very carefully, 
along with the difficulty of training GAN models \cite{goodfellow2014generative}.
Moreover, the first CNN (the target model) takes responsibility of
both characteristic consistifying and low-to-high resolution mapping,
which further increases the model training difficulty dramatically.

In this work, we solve the problem of super-resolving
genuine LR facial images by formulating a 
\textbf{\em Characteristic Regularisation} (CR) method
(Fig \ref{fig:designs}(c)).
In contrast to conventional image SR methods,
we particularly leverage the unpaired 
genuine LR images in order to take into account
their characteristics information for 
facilitating model optimisation.
Unlike cycle consistency based UDA methods,
we instead {\em leverage the artificial LR images
as regularisation target} in order to
separately learn the tasks of characteristic consistifying
and image super-resolution.
Specifically, we perform multi-task learning with the auxiliary task 
as {\em characteristic consistifying} (CC) for transforming genuine LR images
into the artificial LR characteristics,
and the main/target task as {\em image SR} for super-resolving 
both regularised and down-sampled LR images concurrently.
Since there is no HR images coupled with genuine LR images,
we consider to align pixel content in the LR space
by down-sampling the super-resolved images.
%
This avoids the use of cycle consistency
and their learning limitations.
\changed{To make the super-resolved images with good facial identity 
	information, we formulate an unsupervised semantic adaptation
	loss by aligning with the face recognition feature distribution 
	of auxiliary HR images.}

Our CR method can be understood from two perspectives:
(i) As splitting up the whole system into a model for 
image characteristic consistifying 
and a model for image SR. With the former model taking the responsibility 
of solving the characteristic discrepancy, 
the SR model can better focus on learning the resolution enhancement. 
This is in a divide-and-conquer principle.
(ii) As a deeply supervised network \cite{lee2015deeply}, 
providing auxiliary supervision improves accuracy and convergence speed \cite{szegedy2015going}. 
In our case specifically, 
it allows for better and more efficient pre-training of SR module 
using paired artificial LR and HR images, 
pre-training of CC module by genuine and artificial LR images, 
and fast convergence in training the full
CC+SR model.

The \textbf{contributions} of this work are as follows:
(1) We propose a novel super-resolution (SR) method 
for genuine low-resolution facial imagery.  
It combines the advantages of the existing image SR and 
unsupervised domain adaptation methods by a divide-and-conquer strategy.
(2) The proposed {\em Characteristic Regularisation}
enables computationally more tractable model training
and better model generalisation capability.
(3) We introduce a new unsupervised learning loss function
without the limitations of cycle consistency.
We conduct extensive experiments on 
super-resolving both {\em genuine} and {\em artificial} LR facial imagery,
with the former sampled from challenging unconstrained social-media and surveillance videos.
The results validate the superiority of our model over
the state-of-the-art image SR and domain adaptation methods.

\begin{figure*} 
	\center
	\includegraphics[width=.98\linewidth]{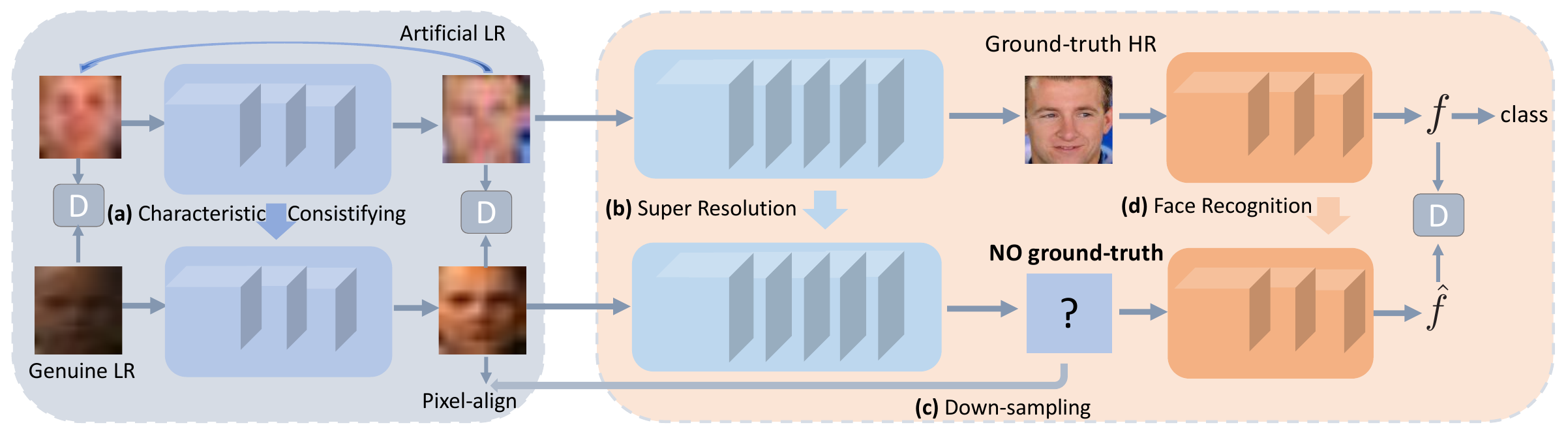}
	\caption{An overview of the proposed {\em Characteristics Regularisation} (CR) approach for super-resolving genuine LR facial imagery data.
		The CR model performs multi-task learning.
		{\bf (a)} The auxiliary task is 
		{\em characteristic consistifying} in order to 
		transform genuine LR images
		into the artificial LR characteristics.
		{\bf (b)} The main task is {\em image SR} allowing for super-resolving 
		both regularised and down-sampled artificial LR images concurrently.
		{\bf (c)} Due to no paired HR images, 
		we propose to align pixel content in the LR space
		by down-sampling the super-resolved images.
		\changed{{\bf (d)} 
			To make the super-resolved images with good facial identity 
			information, we formulate an unsupervised semantic adaptation
			loss term in the adversarial learning spirit,
		 w.r.t. a supervised face recognition model trained on auxiliary HR images.}
	}
	\label{fig:framework}
\end{figure*}

\section{Related Work}
\noindent{\bf Image super-resolution}.
Most state-of-the-art image SR methods 
usually learn the 
resolution mapping functions with 
artificially down-sampled LR and ground-truth HR image pairs \cite{dong2016image,kim2016accurate,lai2017deep,mao2016image,zhang2018super,zhu2016deep}. 
Such pairwise supervised learning becomes infeasible when
there is no HR-LR training pairs, e.g. genuine low-quality 
facial imagery data from in-the-wild social media and surveillance videos.
Recently,
Generative Adversarial Networks (GANs) based image SR models 
\cite{chen2018fsrnet, ledig2017photo, yu2018super}
have been developed. 
They additionally exploit an unsupervised adversarial learning loss
on top of the conventional MSE loss.
These GAN methods often produce more photo-realistic and visually appealing images.
While the GAN loss is unsupervised, these methods still heavily
rely on the paired LR and HR training data therefore 
remaining unsuitable for genuine facial image SR.
Most of the existing works consider mostly
an {\em artificial} image SR problem. The LR images are 
{\em synthesised} by pre-defined down-sampling processes,
with different imaging characteristics to 
{\em genuine} LR images.
When applied to {genuine} LR images, this acquisition difference
causes the domain shift problem for the existing supervised SR models
trained by artificial LR images.
Learning to super-resolve {genuine} LR imagery is much harder due to 
no paired HR ground-truth available for model training.

\noindent{\bf Super-resolution for genuine imagery}.
There are a few recent attempts on resolving genuine
image SR \cite{bulat2017super,bulat2018learn,cheng2018low,cheng2018surveillance,shocher2018zero}.
In particular, 
Shocher et al. \cite{shocher2018zero} learn an image-specific CNN model 
for each test time based on the internal image statistics.
Whilst addressing the problem of pairwise training
data limitation, 
this method is computationally expensive from 
on-the-fly per test image model learning, even with small (compact) neural networks.
Bulat and Tzimiropoulos \cite{bulat2017super} 
develop an end-to-end adversarial learning method for both face SR and alignment
with the main idea that jointly detecting landmarks
provides global structural guidance to the super-resolution process. 
This method is however sensitive to alignment errors.
Bulat et al. \cite{bulat2018learn} utilise the external training pairs, 
where the LR inputs are generated by simulating the real-world image degradation 
instead of simply down-sampling. 
This method presents an effective attempt 
on genuine LR image enhancement.
However, it suffers from an issue of model input discrepancy
between training (simulated genuine LR images) and test
(genuine LR images).
On the other hand, 
unsupervised domain adaptation (UDA) models
\cite{zhu2017unpaired,yi2017dualgan,kim2017learning}
also offer a potential solution for genuine LR image super-resolution.
This approach often uses some cycle consistency based loss function for 
model optimisation, which unfavourably makes the training difficult and ineffective.
\CRC{To tackle the absence of pixel-alignment between LR and HR
	training images, 
Cheng et al.~\cite{cheng2018low,cheng2018surveillance} 
explore 
facial identity information 
to constrain the learning of a SR model.
However, 
this semantic regularisation fails to yield appealing
visual quality.
}

In contrast to all the existing solutions,
we formulate a unified method that enjoy the strengths 
of both conventional SR and UDA methods in
a principled manner.
In particular, we separate the image characteristic consistifying (adaptation)
and image super-resolution tasks by
characteristic regularisation.
Importantly, this makes the model training more effective and 
computationally more tractable,
leading to superior model generalisation capability.

\section{Method}

We aim to obtain a super-resolved HR image $I^\text{sr}$
from an input genuine LR facial image $I^\text{lr}$
with unknown noise characteristics. 
In real-world applications, we have no access to
the corresponding HR counterparts for $I^\text{lr}$.
This prevents the {\em supervised} model training 
of low-to-high resolution mapping between them. 
One solution is to leverage auxiliary HR facial image data
$I^\text{hr}_\text{aux}$.
We first give an overview of existing image SR models before 
introducing the proposed characteristic regularisation method.

\subsection{Facial Image Super-Resolution}
Given auxiliary HR facial images $I^\text{hr}_\text{aux}$,
we can easily generate corresponding LR images $I^\text{lr}_\text{aux}$
by down-sampling.
With such paired data, we can train a 
common supervised image SR CNN model optimised 
by some pixel alignment loss constraint such as the Mean-Squared Error (MSE)
between the resolved and ground-truth images
\cite{ledig2017photo}:
\begin{equation}
\mathcal{L}_\text{sr} =  
\| I^\text{hr}_\text{aux} - \phi_\text{sr}(I^\text{lr}_\text{aux})) \|_2^2
\label{eqn:SR}
\end{equation}
The learned non-linear mapping function $\phi_\text{sr}$ 
can be then applied to 
super-resolve LR test images as:
\begin{equation}
I^\text{sr} =  \phi_\text{sr}(I^\text{lr}_\text{aux})
\label{eqn:resolved}
\end{equation}
This model deployment expects the test data 
with similar distribution as the artificial 
LR training facial images.
If feeding genuine LR images, 
the model may generate much poor results 
due to the domain gap problem.

\subsection{Characteristics Regularisation}

To address the domain gap in SR, 
we take a divide-and-conquer strategy:
first characteristic consistifying, 
then image super-resolving.
Specifically, 
a given genuine LR image is first transformed
into that with similar appearance characteristics
as artificial LR images.
Then, the SR model is able to better perform 
image super-resolving.
To that end, we exploit the unsupervised GAN learning framework \cite{goodfellow2014generative}.
The objective is to learn a model that can 
synthesise facial images indistinguishable from artificial LR data
with condition on genuine LR input. 

Formally, 
the Characteristics Regularisation (CR) GAN model 
consists of a discriminator $D$ that is optimised to distinguish 
whether the input is an artificial down-sampled LR or not, 
and a characteristics regularisor $\phi_\text{cr}$ that 
transforms a genuine LR input $I^\text{lr}$ 
to fool the discriminator to classify the transformed $\phi_\text{cr}(I^\text{lr})$
as an artificial image. 
The objective function can be written as:
\begin{equation}
\begin{split}
\mathcal{L}_\text{gan} = \; & \mathbb{E}_{I^\text{lr}_\text{aux}}[\log D(I^\text{lr}_\text{aux})] + \\ 
\; & \mathbb{E}_{I^\text{lr}}[\log\big(1-D(\phi_\text{cr}(I^\text{lr}))],
\end{split}
\label{eq:da}
\end{equation}
where the characteristics regularisor $\phi_\text{cr}$ 
tries to minimise the objective value
against an adversarial discriminator D that instead tries to
maximise the value. 
The optimal adaptation solution is obtained as:
\begin{equation}
G^* = \arg \; \min_{\phi_\text{cr}} \; \max_D \mathcal{L}_\text{gan}.
\label{eq:GAN_opt}
\end{equation}


\changed{To better connect the characteristics regularisation $\phi_\text{cr}$ with the super-resolving $\phi_\text{sr}$ module,
we enable an end-to-end training for the auxiliary artificial LR branch by
additionally learning a mapping from the down-sampled artificial LR images
to the transformed pseudo genuine LR counterparts.
More specifically,
we first generate \textit{pseudo} genuine LR images
by an inverse process of CR,
i.e. transforming an artificial LR image to fool the discriminator to classify the transformed $\tilde{\phi}_\text{cr}(I^\text{lr}_\text{aux})$
as a genuine LR image:
\begin{equation}
\begin{split}
\arg \; \min_{\tilde{\phi}_\text{cr}} \; \max_{\tilde{D}} \; & \mathbb{E}_{I^\text{lr}}[\log \tilde{D}(I^\text{lr})] + \\ 
\; & \mathbb{E}_{I^\text{lr}_\text{aux}}[\log\big(1-\tilde{D}(\tilde{\phi}_\text{cr}(I^\text{lr}_\text{aux}))],
\end{split}
\label{eq:ida}
\end{equation}
This is learned independently. 
Then, $\phi_\text{cr}$ can be jointly optimised by a loss formula as:
\begin{equation}
\mathcal{L_\text{cr}} = 
|| I^\text{lr}_\text{aux}-\phi_\text{cr}(\tilde{\phi}_\text{cr}(I^\text{lr}_\text{aux})) ||_2^2 +
\lambda  \mathcal{L}_\text{gan}
\label{eq:cr_loss}
\end{equation}
where $\lambda$ is a weight hyper-parameter.
We set $\lambda=0.2$ in our experiment.
We found this design improves 
the stability of end-to-end joint training for $\phi_\text{cr}$ and $\phi_\text{sr}$.
}

\subsection{Super-Resolving Regulated Images}

If the CR module is perfect in characteristic consistifying,
the SR module $\phi_\text{sr}$ trained on the auxiliary facial data can be 
directly applied.
However, this is often not the truth in reality.
So, it is helpful to further fine-tune $\phi_\text{sr}$
on the regulated data $\phi_\text{cr}(I^\text{lr})$. 
To do this, we need to address the problem of lacking
HR supervision.
Instead of leveraging the conventional cycle consistency idea,
we adopt a simple but effective pixel-wise distance constraint. 
The intuition is that, a good super-resolved image output, after down-sampling,
should be close to the LR input.
By applying this cheap condition, we do not need 
to access the unknown HR ground-truth.
Formally, we design this SR loss function
for regulated LR images as:
\begin{equation}
\mathcal{L_\text{cr-sr}} = 
|| f_\text{DS}\Big(\phi_\text{sr}\big(\phi_\text{cr}(I^\text{lr})\big) \Big) - \phi_\text{cr}(I^\text{lr}) ||_2^2 
\label{eq:contents_constraint}
\end{equation}
where $f_\text{DS}$ refers to the down-sampling function.


\subsection{\changed{Unsupervised Semantic Adaptation}}
\changed{
Apart from visual fidelity,
the SR output is also required to be semantically
meaningful with good identity information.
To this end, we form an unsupervised 
semantic adaptation loss term in the adversarial learning spirit. 	
The idea is to constrain the perceptual {\em feature} distribution 
of super-resolved facial images by matching 
the feature statistics of auxiliary HR images $I_\text{aux}^\text{hr}$.
It is formally written as:
\begin{equation}
\begin{split}
 & \mathcal{L_\text{cr-gan}} = \; \mathbb{E}_{I^\text{hr}_\text{aux}}[\log D'(\phi_\text{fr}(I^\text{hr}_\text{aux}))] \; + \\ 
 \; & \mathbb{E}_{\phi_\text{sr}(\phi_\text{cr}(I^\text{lr}))}\Big[\log\Big(1-D'\big(\phi_\text{fr}\big(\phi_\text{sr}(\phi_\text{cr}(I^\text{lr}))\big)\big)\Big)\Big]
\end{split}
\label{eq:fidelity_constraint}
\end{equation}
where $\phi_\text{fr}$ is a CentreFace~\cite{wen2016discriminative} based 
feature extractor pre-trained with
$I_\text{aux}^\text{hr}$.
This loss is unsupervised without the need for identity labels of
{\em genuine} LR training images.
Compared to image based GAN loss,
it is found more efficient and easier to train
in a low-dimension feature space.
}

\begin{figure*} 
	\centering
	\includegraphics[width=.6\linewidth]{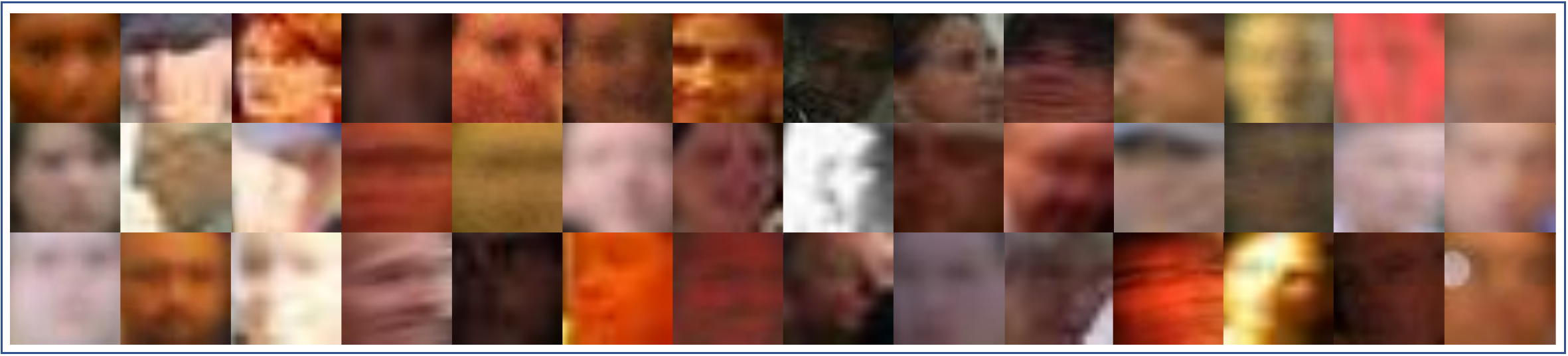}	\includegraphics[width=.6\linewidth]{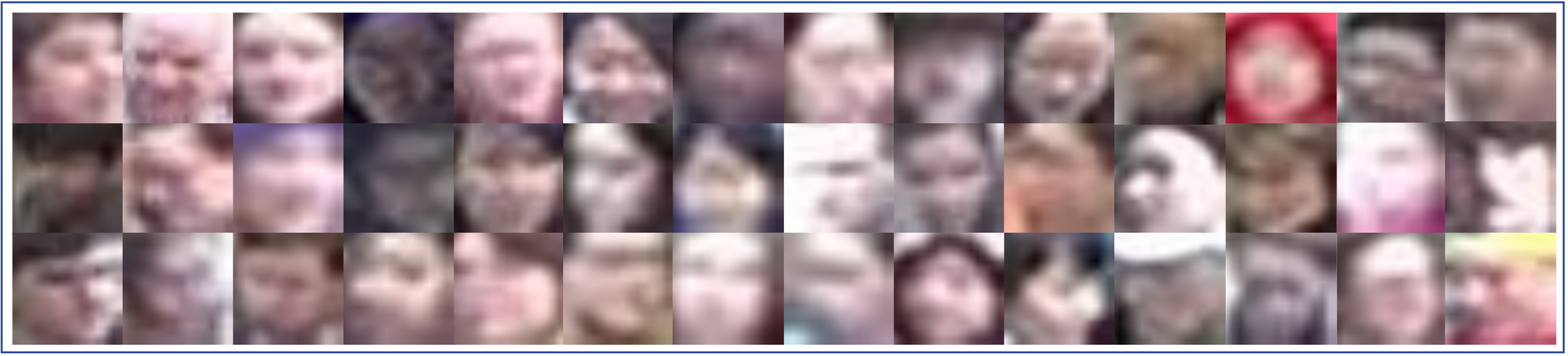}
	\caption{Examples of genuine facial images randomly sampled from LR-PIPA
		({\bf top}) and LR-DukeMTMC ({\bf bottom}).} 
	\label{fig:dataset}
\end{figure*}

\subsection{Model Training and Inference}  

Due to introduction of characteristic regularisation in the middle of our full model,
more effective model training is enabled. 
It facilitates a two-staged training strategy. 
In the first stage, we pre-train the CNN for image SR 
on the auxiliary LR-HR paired facial data,
\changed{the CentreFace model on HR images}, and
the CNN for characteristic regularisation \changed{and the inverse CR}
on unpaired genuine and artificial LR images
in parallel.
In the second stage, the cascaded CR and SR CNNs are fine-tuned together
on all the training data.

\vspace{0.1cm}
\noindent {\bf CNN for image super-resolution.}
We train the image SR model $\phi_\text{sr}$ as \cite{ledig2017photo}
by deploying the pixel-wise MSE loss function
(Eq \eqref{eqn:SR}). This model training 
benefits from the normal adversarial loss for 
achieving better perceptual quality.
Other existing image SR methods \cite{kim2016accurate,dong2016image} can be readily considered
in our framework.

\vspace{0.1cm}
\noindent {\bf CNN for characteristic regularisation.}
We train the CNN for characteristic regularisation $\phi_\text{cr}$
by an adversarial loss and a pixel-wise loss jointly
(Eq \eqref{eq:cr_loss}).

\vspace{0.1cm}
\noindent {\bf Full model.}
In the second stage, we further fine-tune 
both CNN models $\phi_\text{sr}$ and $\phi_\text{cr}$ jointly.
The overall objective loss for training the full model is
formulated as:
\begin{equation}
\mathcal{L} =
\mathcal{L}_\text{sr}+ 
\lambda_\text{cr}\mathcal{L}_\text{cr}+
\lambda_\text{cr-sr}\mathcal{L}_\text{cr-sr}+\lambda_\text{cr-gan}\mathcal{L}_\text{cr-gan}
\label{eq:overall}
\end{equation}
where $\lambda_\text{cr},\lambda_\text{cr-sr},\lambda_\text{cr-gan}$ 
are the weight parameters of the corresponding loss terms. 
In our experiment, we set 
$\lambda_\text{cr} = 0.06$,
$\lambda_\text{cr-sr} = 0.01$,
$\lambda_\text{cr-gan} = 0.03$
by cross-validation. 

\vspace{0.1cm}
\noindent {\bf Model inference.}
Once trained, we deploy the full model for test,
taking a genuine LR facial image as input, outputting a HR image.

\section{Experiments}
\noindent{\bf Datasets}.
For model performance evaluation, 
we created two new real-world genuine LR facial image datasets sampled from  
web social-media imagery and surveillance videos. 
Following \cite{bulat2018learn},
we define LR faces as those with an average size of $\leq$16$\times$16 pixels.
In particular, we constructed the web social-media based 
real-world face image dataset 
by assembling LR faces 
from the People In Photo Albums (PIPA) benchmark~\cite{piper}, called \textbf{\em LR-PIPA}.
Similarly, we collected LR face images (small faces) from a multi-target multi-camera tracking benchmark 
DukeMTMC~\cite{ristani2014tracking} and built our surveillance video real-world face
dataset, called \textbf{\em LR-DukeMTMC}.
There are 8,641 and 7,044 face images in LR-PIPA and LR-DukeMTMC, respectively.
All the face images were obtained
by deploying the TinyFace detector \cite{hu2017finding}.
We manually filtered out non-face images.
These two new datasets consist of genuinely
real-world LR facial images captured from unconstrained camera views
under a large range of different viewing conditions such as expression, pose, illumination,
and background clutter. We will release both datasets publicly.
We show some randomly selected examples in Fig \ref{fig:dataset}.

\vspace{0.1cm}
\noindent{\bf Training and test data}.
To effectively train a competing model, we need both real-world genuine LR images
and web auxiliary HR facial images.
For the former, we used 153,440 LR face images collected from the
Wider Face benchmark \cite{yang2016wider}. 
This dataset offers rich facial images from a wide variety of social events,
with a high degree of variability in scale, pose, lighting, and background.
For the latter, we selected the standard CelebA benchmark with 202,599 HR web facial images \cite{celebface}.
Such a training set design ensures that each model can be trained with sufficiently diverse
data to minimise the learning bias.
For model test, we utilised the entire LR-PIPA and LR-DukeMTMC.
Both datasets present significant test challenges, as
they were drawn from unconstrained and independent data sources
with arbitrary and unknown noise.

\vspace{0.1cm}
\noindent{\bf Performance evaluation metrics}.
Due to that there are {\em no} ground-truth 
HR data of \textit{genuine} LR facial images,
it is impossible to conduct pixel based performance evaluation
and comparison.
We utilise the Frechet Inception Distance (\textbf{\em FID})
\cite{heusel2017gans} 
to assess the quality of resolved face images,
similar to the state-of-the-art method \cite{bulat2018learn}. 
Specifically, \textbf{\em FID} is measured by 
the Frechet Distance between two multivariate Gaussian distributions.
%


\noindent{\bf Implementation details}.
We performed all the following experiments in Tensorflow.
We used the residual blocks \cite{he2016deep} as the backbone unit of
our network.
In particular, we used 3 residual blocks in 
the net for the characteristics regularisation module $\phi_\text{cr}$ and $\tilde{\phi}_\text{cr}$,
and we further adapted the SRGAN (3 groups containing 
12/3/2 residual blocks, respectively. 
Resolution is increased 2 times across each group) \cite{ledig2017photo} 
for our facial SR module $\phi_\text{sr}$.
The adversarial discriminator for $\phi_\text{cr}$ and $\tilde{\phi}_\text{cr}$ both consist of  
6 residual blocks,
followed by a fully connected layer.
The adversarial discriminator $D'$ for semantic adaptation consists of 5 fully connected layers. 
All LR images were sized at $16\times16$.
The scale of real-world facial image super-resolution was 16 (4$\times$4) times,
i.e. the output size is $64\times64$.
We set 
the learning rate to $10^{-4}$,
the batch size to $16$.
The SR module ($\phi_{sr}$ in Fig.~\ref{fig:framework}) 
was pre-trained on CelebA face dataset with down-sampled 
artificial LR and HR image pairs for 100 epochs. 
And the characteristic consistifying module was trained with 
unpaired genuine and artificial LR images (down-sampled from CelebA dataset)
for 130 epochs.
The end-to-end full model was jointly trained by 10 epochs.

\begin{table} [!h]
	\centering
	\setlength{\tabcolsep}{0.2cm}
	\begin{tabular}{c||c|c}
		\hline
		Dataset & LR-PIPA & LR-DukeMTMC \\
		\hline \hline
		VDSR~\cite{kim2016accurate} & 94.49 & 229.56 \\
		SRGAN~\cite{ledig2017photo} & 103.85 & 232.38 \\
		FSRNet~\cite{chen2018fsrnet} & 117.19 & 218.30 \\
		SICNN~\cite{zhang2018super} & 129.23 & 223.08 \\
		\hline
		CycleGAN~\cite{zhu2017unpaired} & 33.62 & 42.41 \\ 
		\CRC{CSRI~\cite{cheng2018low}} & 104.68 & 240.99 \\
		LRGAN~\cite{bulat2018learn} & 29.80 & 31.20 \\ 
		\hline 
		\bf CR (Ours) & \bf 23.09 & \bf 25.56  \\  
		\hline
	\end{tabular}
	\vskip 0.1cm
	\caption{
		Comparing the image quality on {\em genuine} LR
		facial image super-resolution. Metric: FID. \textbf{Lower is
			better.}
	}
	\label{table:fid}
\end{table}

\subsection{Test Genuine Low-Resolution Facial Images}
\noindent \textbf{\em Competitors}.
To evaluate the effectiveness of our CR model 
for genuine facial image SR, we compared with four {\em groups} of 
the state-of-the-art methods including, 
two generic image SR models (VDSR~\cite{kim2016accurate}, SRGAN~\cite{ledig2017photo}),
one image-to-image translation model (CycleGAN \cite{zhu2017unpaired}),
one {\em non}-genuine face SR model (FSRNet~\cite{chen2018fsrnet}), 
one \CRC{UDA-based} genuine face SR model (LRGAN \cite{bulat2018learn}),
\CRC{and one facial identity-guided genuine SR model (CSRI~\cite{cheng2018low}).
Same as our CR, CycleGAN, LRGAN and CSRI were trained using genuine LR images,
while the others with artificial LR only as they 
need
pixel-aligned LR and HR training image pairs.
}

\vspace{0.1cm}
\noindent{\bf Results}.
The results of these methods are compared in Table~\ref{table:fid}.
We have the following observations:
%
{\bf (1)} The proposed CR model achieves the best FID score among all the competitors,
suggesting the overall performance advantage of our approach 
on super-resolving genuine LR facial images.
%
{\bf (2)} Generic image SR methods (VDSR, SRGAN) perform the worst, as expected,
although re-trained by the large-scale CelebA face data 
with artificial LR and HR image pairs. 
This is due to the big image characteristics difference between 
the source artificial LR and the target genuine LR images. 
%
{\bf (3)}
By considering the problem from
image-to-image domain adaptation perspective, CycleGAN is shown to be
superior than VDSR and SRGAN models.
This is because of no domain gap problem.
However, it is less optimal than modelling explicitly genuine LR face images in the SR
process, as compared to the two specifically designed genuine LR facial
image super-resolution models, CR and LRGAN. This is more so
in surveillance videos (LR-DukeMTMC).
%
{\bf (4)}
\CRC{With the high-level facial identity constraint,
	CSRI cannot achieve satisfactory low-level visual fidelity in the pixel space.}
%
{\bf (5)}
Despite modelling facial prior explicitly,
FSRGAN fails to improve meaningfully over generic SR methods (VDSR, SRGAN). 
This is due to the significant domain gap between the genuine and
artificial LR facial images, leading to difficulty in
inferring useful facial content and structural prior from the
low-quality genuine LR images.
%
{\bf (6)}
As a state-of-the-art model, LRGAN demonstrates its advantages over other models by learning
explicitly the image degradation process. 
However, it is clearly outperformed by the proposed CR model.
This suggests the overall performance advantages of our method.


\begin{figure} 
	\centering
	\includegraphics[width=0.92\linewidth]{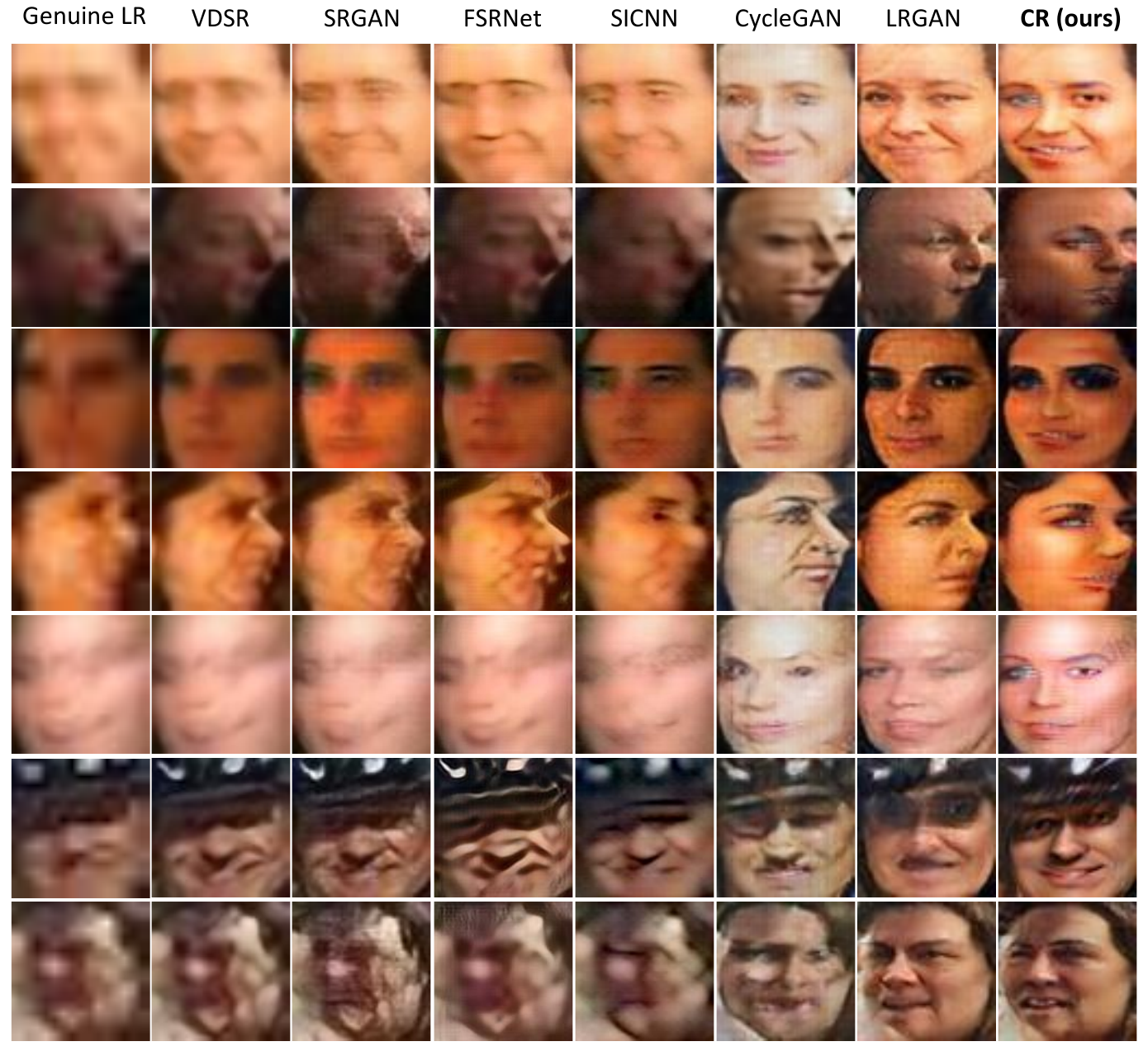}
	\includegraphics[width=0.921\linewidth]{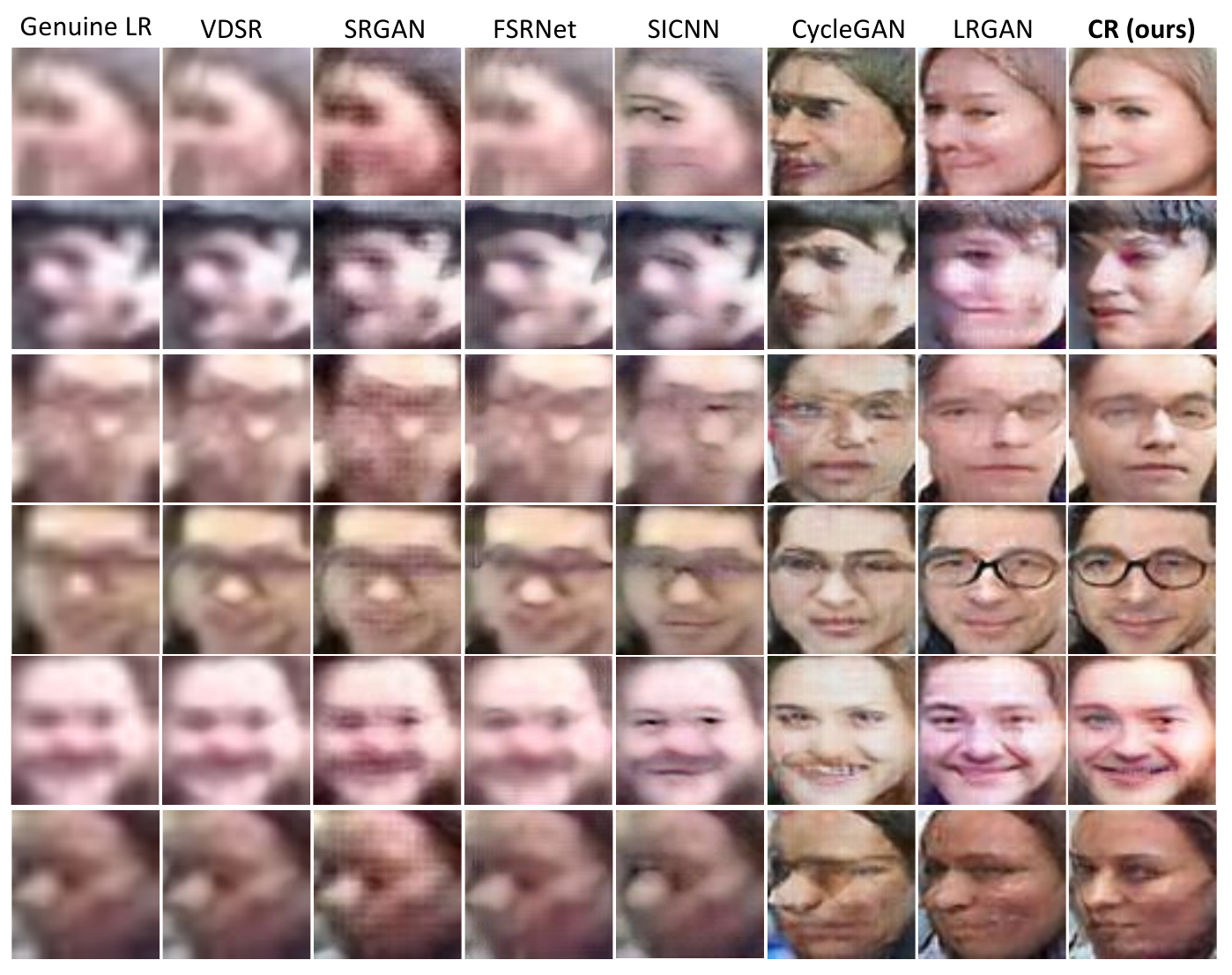}
	\caption{
		Examples of genuine LR image super-resolution on 
		({\bf top}) LR-PIPA and ({\bf bottom}) LR-DukeMTMC.
	} 
	\label{fig:sr_qualitative}
\end{figure}

\vspace{0.1cm}
\noindent {\bf Qualitative evaluation}.
To conduct visual comparisons between different alternative methods,
we provided SR results of random genuine LR facial images in Fig \ref{fig:sr_qualitative} .
Overall, the visual examination is largely consistent 
with the numerical evaluation.
Specifically, existing methods
tend to generate images with severe blurry and artefact either globally (VDSR, SRGAN)
or locally (CycleGAN, LRGAN).
In contrast, CR can yield 
HR facial images with much better fidelity
in most cases.
This visually verifies the superiority of our method in
super-resolving genuine LR facial images. 

\vspace{0.1cm}
\noindent {\bf Model complexity}.
We compared the top-3 models 
(CR, LRGAN \cite{bulat2018learn}, and CycleGAN \cite{zhu2017unpaired})
in three aspects:
(1) Model parameters:
2.7, 4.0, and 21 million;
(2) Training time: 46, 72, and 81 hours; and 
(3) Per-image inference time: 7.5, 6.6, and 150 ms,
using a Tesla P100 GPU.
Therefore, our model is the most compact and
most efficient.


\vspace{-0.1cm}
\subsection{\changed{Face Recognition on Genuine LR Face Imagery}}
\changed{
We tested the benefit of image SR on a downstream task, {\em face recognition},
on the LR-PIPA dataset.
We used the CentreFace model trained on the auxiliary HR images
and the CMC rank metrics. 
The results in Table~\ref{table:fr} show that:
(1) Directly using raw LR images leads to very poor recognition rate,
due to lacking fine-grained facial trait details.
(2) CR achieves the best performance gain as compared to all the strong
competitors.
(3) Interestingly, LRGAN gives a negative recognition margin,
mainly due to introducing more identity-irrelevant enhancement
despite good visual fidelity. 
}

\vspace{-0.2cm}
\begin{table} [!h]
	\centering
	\setlength{\tabcolsep}{0.9cm}
	\begin{tabular}{c||c}
		\hline
		Dataset & Rank-1 (\%)  \\
		\hline \hline
		VDSR~\cite{kim2016accurate} & 25.45  \\
		SRGAN~\cite{ledig2017photo} & 27.00  \\ 
		FSRNet~\cite{chen2018fsrnet} & 26.50 \\ 
	        SICNN~\cite{zhang2018super} & 28.85 \\
	        \hline
		CycleGAN~\cite{zhu2017unpaired} & 25.12  \\  
		\CRC{CSRI~\cite{cheng2018low}} & 29.59 \\
		LRGAN~\cite{bulat2018learn} & 21.99  \\  
		\hline 
		\bf CR (Ours) & \bf 30.53   \\ 
		\hline 
		\hline
		\em Raw LR input & 24.83 \\ 
		\hline
	\end{tabular}
	\vskip 0.1cm
	\caption{
		Face recognition performance
		on super-resolved {\em genuine} LR
		images from LR-PIPA.
		Metric: Rank-1. 
		\textbf{Higher is better.}
	}
	\label{table:fr}
\end{table}

\subsection{Test Artificial Low-Resolution Facial Images}

\vspace{-0.2cm}
\begin{table} [!h]
	\centering
	\setlength{\tabcolsep}{0.5cm}
	\begin{tabular}{c||c|c}
		\hline
		Metric & PSNR & SSIM \\
		\hline 
		VDSR~\cite{kim2016accurate} & \bf 26.31 & 0.7918 \\
		SRGAN~\cite{ledig2017photo} & 25.10 & 0.7873 \\
		FSRNet~\cite{chen2018fsrnet} & 25.10 & 0.7234 \\
		SICNN~\cite{zhang2018super} & 26.10 & 0.7986 \\
		\hline
		CycleGAN~\cite{zhu2017unpaired} & 18.85 & 0.6061 \\ 
		\CRC{CSRI~\cite{cheng2018low}} & 25.40 & 0.7388 \\
		LRGAN~\cite{bulat2018learn} & 21.88 & 0.6869 \\ 
		\hline 
		\bf CR (Ours) & 25.50 & \bf 0.8184  \\
		\hline
	\end{tabular}
	\vskip 0.1cm
	\caption{
		Comparison of state-of-the-art methods on {\em artificial} LR
		facial image super-resolution. Dataset: Helen. 
		Metric: PSNR \& SSIM. 
		\textbf{Higher is better.}
	}
	\label{table:sr_ideal}
\end{table}

For completeness, we tested model performance
in artificial LR facial images as 
in conventional SR setting.

\vspace{0.1cm}
\noindent {\bf Model deployment}.
By design, our CR model is trained for super-resolving genuine LR facial imagery.
However, it can be flexibly deployed {\em without} 
the characteristic regulation module, when artificial LR test images are given.

\vspace{0.1cm}
\noindent {\bf Dataset}.
In this evaluation we selected the Helen face dataset \cite{le2012interactive} 
with 2,330 images.
We produced the artificial LR test images by bicubic down-sampling, \changed{as the conventional SR evaluation setting}.

\vspace{0.1cm}
\noindent {\bf Metrics}.
For performance evaluation,
we used the common Peak Signal-to-Noise Ratio (PSNR) 
and structure similarity index (SSIM) \cite{wang2004image}. 
This is because, we have the ground-truth HR images
for pixel-level assessment in this case.

\vspace{0.1cm}
\noindent {\bf Results}.
Table~\ref{table:sr_ideal} compares the performances on normal Helen LR facial images 
of our CR and state-of-the-art SR methods.
It is observed that our method can generate better results
than all the competitors except VDSR for the PSNR metric.
Interestingly, CR outperforms SRGAN which is actually our SR module.
This implies that the model generalisation for conventional SR tasks
can be improved by 
the proposed unsupervised SR learning objective
(Eq \eqref{eq:contents_constraint}).




\subsection{Component Analysis and Discussion}

We conducted a series of model component analysis
for giving insights to our model performance.

\vspace{-0.1cm}
\begin{table} [!h]
	\centering
	\setlength{\tabcolsep}{0.5cm}
	\begin{tabular}{c||c|c}
		\hline
		Dataset & LR-PIPA & LR-DukeMTMC \\
		\hline \hline
		W/O CR & 133.30 & 190.73 \\ 
		\hline
		W/ CR & \bf 23.09 & \bf 25.56 \\
		\hline
	\end{tabular}
	\vskip 0.1cm
	\caption{Effect of characteristics regularisation (CR). 
		Metric: FID. 
	}
	\label{table:CR}
\end{table}

\vspace{0.1cm}
\noindent {\bf Characteristic regularisation}.
We evaluated the effect and benefits of characteristic regularisation (CR)
on model performance. 
We compared with a {\em baseline} which learns the SR module from
genuine and artificial LR images {\em jointly}.
The baseline model needs to fit heterogeneous input data distributions.
The training loss function is 
$
\mathcal{L}_\text{base} =
\mathcal{L}_\text{sr}+
\lambda_\text{cr-sr}\mathcal{L}_\text{cr-sr}+
\lambda_\text{cr-gan}\mathcal{L}_\text{cr-gan}
$.
This allows for testing the exact influence of
characteristic consistifying.
Table \ref{table:CR} shows that 
CR plays a key role 
for enabling the model to super-resolve genuine LR facial images.
Without CR, the model fails to properly accommodate the genuine data,
partly due to an extreme modelling difficulty for
learning such a cross-characteristics cross-resolution mapping


\begin{table} [!h]
	\centering
	\setlength{\tabcolsep}{0.3cm}
	\begin{tabular}{c||c|c}
		\hline
		Dataset & LR-PIPA & LR-DukeMTMC \\
		\hline \hline
		FID(G-LR, A-LR) & 40.72 & 86.23 \\ 
		\hline
		FID(R-LR, A-LR) & \bf 19.49 & \bf 24.32 \\ 
		\hline
	\end{tabular}
	\vskip 0.1cm
	\caption{Evaluate regulated LR images (R-LR).
	G-LR: Genuine LR images; A-LR: Artificial LR images. 
	}
	\label{table:CR_result}
\end{table}

We further examined the result of characteristic consistifying,
i.e. the regulated LR images.
To this end, we measured the FID between artificial and regulated LR images,
in comparison to that between artificial and genuine LR images.
Table \ref{table:CR_result} shows that
although regulated LR images match significantly better to artificial LR data
than their genuine counterparts,
the distribution difference remains.
This suggests the necessity of fine-tuning the SR module 
on the regulated LR images (the second training stage).

\begin{figure} [!h]
	\centering
	\includegraphics[width=.95\linewidth]{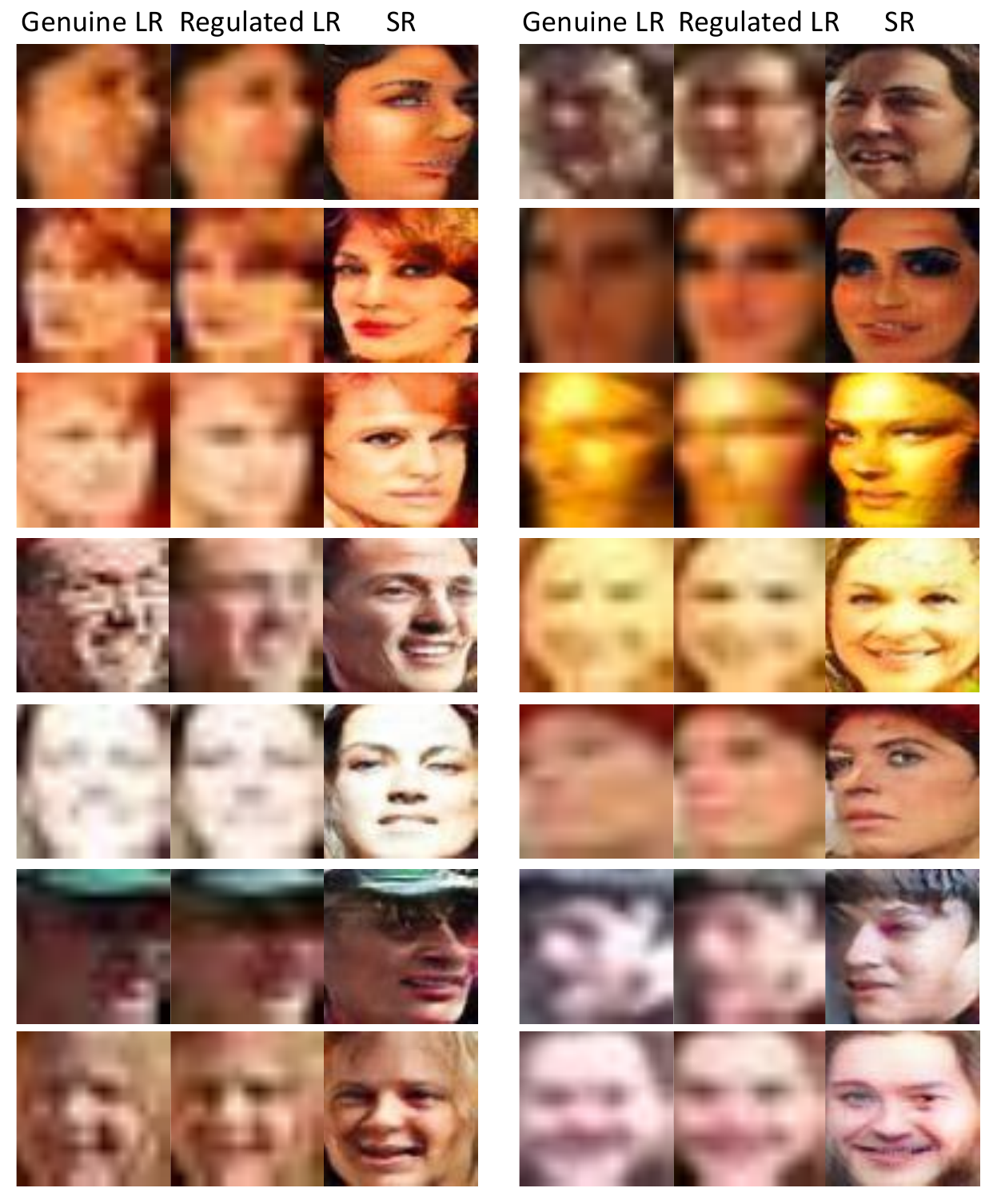}
	\caption{
		Genuine LR {\em vs.} regulated LR {\em vs.} resolved face images.
	} 
\label{fig:lr_adaptation}
\end{figure}

We showed qualitative results Fig.~\ref{fig:lr_adaptation}.
It is observed that compared to the genuine LR input, 
the regulated images have clearer contour of facial components, 
better lighting conditions and less blur,
i.e. much closer to artificial LR data.
This eases the subsequent SR job.





\begin{table} [!h]
	\centering
	\setlength{\tabcolsep}{0.3cm}
	\begin{tabular}{c||c|c}
		\hline
		Dataset & LR-PIPA & LR-DukeMTMC \\
		\hline \hline
		W/O SR-RI & 111.01 & 113.70 \\ 
		\hline
		W/ SR-RI & \bf 23.09 & \bf  25.56 \\
		\hline
	\end{tabular}
	\vskip 0.1cm
	\caption{Effect of super-resolving regulated images (SR-RI). Metric: FID. \textbf{Lower is better.} 
	}
	\label{table:lr_srri}
\end{table}

\vspace{0.1cm}
\noindent {\bf Super-resolving regulated images}.
In the second training stage, we fine-tune the SR module
for better super-resolving regulated LR images.
We evaluated the effect of this design.
Table \ref{table:lr_srri} shows that 
the model performance drops noticeably
without the proposed SR model fine-tuning
on regulated LR images.
This is consistent with the observation in Table \ref{table:CR_result}.

\begin{table} [!h]
	\centering
	\setlength{\tabcolsep}{0.3cm}
	\begin{tabular}{c||c|c}
		\hline
		Dataset & LR-PIPA & LR-DukeMTMC \\
		\hline \hline
		W/O UL & 25.30 & 26.11 \\ 
		\hline
		W/ UL & \bf 23.09 & \bf 25.56 \\
		\hline
	\end{tabular}
	\vskip 0.1cm
	\caption{Effect of unsupervised loss (UL) for super-resolving regulated images. Metric: FID. \textbf{Lower is better.} 
	}
	\label{table:loss_srri}
\end{table}

\begin{figure} [!h]
	\centering
	\includegraphics[width=.95\linewidth]{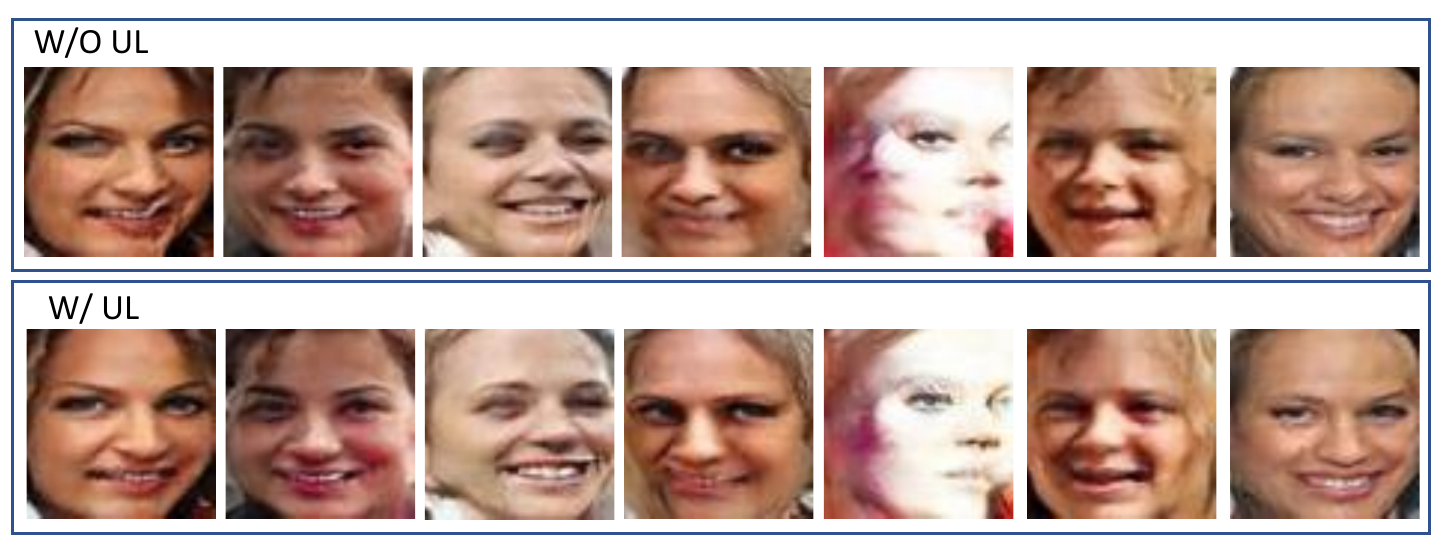}
	\caption{
		Visual examination: W/O {\em vs.} W/ unsupervised loss (UL) 
		in super-resolving regulated images.
	} 
	\label{fig:loss_srri_ex}
\end{figure}

Recall that we introduce an unsupervised SR loss (Eq \eqref{eq:contents_constraint}) for regulated LR images due to no HR ground-truth.
We consider pixel-wise alignment in LR image space,
without the need for cycle consistency.
We tested its impact on the model performance.
Table \ref{table:loss_srri} shows that
applying this loss can clearly boost the 
fidelity quality of resolved faces.
Also, we found that it makes the model training more stable.
Further qualitative evaluation in
Fig \ref{fig:loss_srri_ex} shows that
the unsupervised SR loss
can help reduce the noise and distortion in SR,
leading to visually more appealing results.

%

\section{Conclusion}
We present a Characteristic Regularised (CR) method
for super-resolving genuine LR facial imagery.
This differs from most SR studies focusing on 
artificial LR images with limited model generalisation
on genuine LR data
and UDA methods suffering ineffective training.
In comparison, CR possesses the modelling merits of 
previous SR and UDA models end-to-end,
solves both domain shift and ineffective model training, 
and simultaneously takes advantage
of rich resolution information from abundant auxiliary training data. 
We conduct extensive comparative experiments 
on both genuine and artificial LR facial images.
The results show the performance and generalisation advantages of our method
over a variety of state-of-the-art image SR and UDA models.
We carry out detailed model component analysis for
revealing the model formulation insights.

\section*{Acknowledgement}
\CRC{
This work was partially supported by the Alan Turing Institute
Turing Fellowship, the Innovate UK Industrial Challenge Project on Developing and Commercialising Intelligent Video Analytics Solutions for Public Safety (98111-571149), Vision Semantics Ltd, and SeeQuestor Ltd.}

{\small
\bibliographystyle{ieee}
\bibliography{egbib}
}

\end{document}